\documentclass[runningheads]{llncs}

\usepackage{pdflscape}
\usepackage{amsmath}
\usepackage{amsfonts}

\usepackage{siunitx}

\usepackage{cite}

  \usepackage[pdftex]{graphicx}
   \DeclareGraphicsExtensions{.pdf,.jpeg,.png}
\usepackage{algorithmicx}
\usepackage[noend]{algpseudocode}
\usepackage{algorithm}

\usepackage{array}
\usepackage{tabularx} 

\begin{document}
\title {A Scalable FPGA-based Architecture for Depth Estimation in SLAM}
%

\author{Konstantinos Boikos \orcidID{0000-0002-4482-906X} \and Christos-Savvas Bouganis }

\institute{Department of Electrical and Electronic Engineering, Imperial College London}

\maketitle              

\email{\{k.boikos14, christos-savvas.bouganis\}@imperial.ac.uk}


\begin{abstract}
The current state of the art of Simultaneous Localisation and Mapping, or SLAM, on low power embedded systems is about sparse localisation and mapping with low resolution results in the name of efficiency. Meanwhile, research in this field has provided many advances for information rich processing and semantic understanding, combined with high computational requirements for real-time processing. This work provides a solution to bridging this gap, in the form of a scalable SLAM-specific architecture for depth estimation for direct semi-dense SLAM. Targeting an off-the-shelf FPGA-SoC this accelerator architecture achieves a rate of more than 60 mapped frames/sec at a resolution of 640x480 achieving performance on par to a highly-optimised parallel implementation on a high-end desktop CPU with an order of magnitude improved power consumption. Furthermore, the developed architecture is combined with our previous work for the task of tracking, to form the first complete accelerator for semi-dense SLAM on FPGAs, establishing the state of the art in the area of embedded low-power systems.
\end{abstract}

\section{Introduction}

In recent years, there has been a lot of interest and research effort surrounding intelligent machines and systems. One area of particular interest is the push towards fully autonomous machines that can move and interact in an unknown environment. This includes emerging applications such as household robots, environment-aware industrial robots, autonomous drones that can operate indoors and self-driving cars among others. One of the core elements in this effort is a family of algorithms and systems called Simultaneous localisation and Mapping (SLAM), which aims to provide a solution to the problem of exploring an unknown environment while keeping tracking of one's own position in it.

From this point, the paper focuses on real-time SLAM, which refers to performing all processing at the camera's rate of operation. The exact rate necessary can vary per application. Focusing on robotics which is one of the central motivations for this work, research has shown that effective localisation needs a performance of at least 30 frames/sec for most moving robotic platforms. Moving to faster platforms, such as self-driving cars and quadcopters, higher framerates are required for SLAM not to fail under agile movement \cite{handa2012real}. Meanwhile, the resolutions used are normally in the region of 640x480, sometimes going up to 960x720. It was found that increasing the resolution provides a reduced benefit to some algorithms \cite{engel2016monodataset}, while the runtime usually increases at least linearly with the number of pixels. However, the state of the art in algorithms has focused significantly on resolutions in this region and research results going down to centimetre level accuracy with a VGA-resolution camera (e.g. \cite{mur2015orb}) seem to indicate that the camera resolution is not currently the limiting factor.

SLAM in the literature is usually comprised of two main tasks \cite{engel14eccv, mur2015orb}. Localisation, often referred to as tracking, is the act of continuously estimating the position and orientation, or pose, of the camera. Mapping is the task of generating and continuously updating a coherent model of the environment based on the sensor observations. These two tasks are very closely interconnected and strongly dependent on each other. Tracking compares the incoming data from the sensor with the map that has been generated to estimate a current pose. Then, the accuracy of that estimation will determine the quality of the updated map, and how close it will be to reality.

In the past, different sensors have been used \cite{fuentes2015visual, lu2009slam} including Lidar, sonar and recently RGB-D cameras e.g. \cite{whelan2015real}. The first two sensors deal with a map usually limited to two dimensions around a moving platform, and were used in early works for their simplicity and effectiveness. They are also  used as complementary sensor-fusion algorithms combined with visual sensors, inertial sensors or both, for applications which demand a high-level of robustness and accuracy, such as self-driving cars. However they are usually heavy, require high power consumption and are mostly constrained in two dimensions, making them unsuitable for many applications, especially as the sole sensory input. Active camera sensors in the RGB-D category recover depth directly by projecting a light pattern in infrared or using time-of-flight. They have enabled high-quality dense 3D reconstruction in indoor spaces \cite{whelan2016elasticfusion} but are constrained in their area of operation because of their design. They are also more expensive and power-hungry than a simple visual sensor, making them less attractive for embedded low-power robotics and outdoor spaces. As such, this work focuses on enabling high quality embedded SLAM using visual information from RGB or greyscale cameras.

\begin{figure}[!ht] 
	\centering
	\includegraphics[trim={1cm 3cm 1cm 5.2cm},clip ,width=1\linewidth]{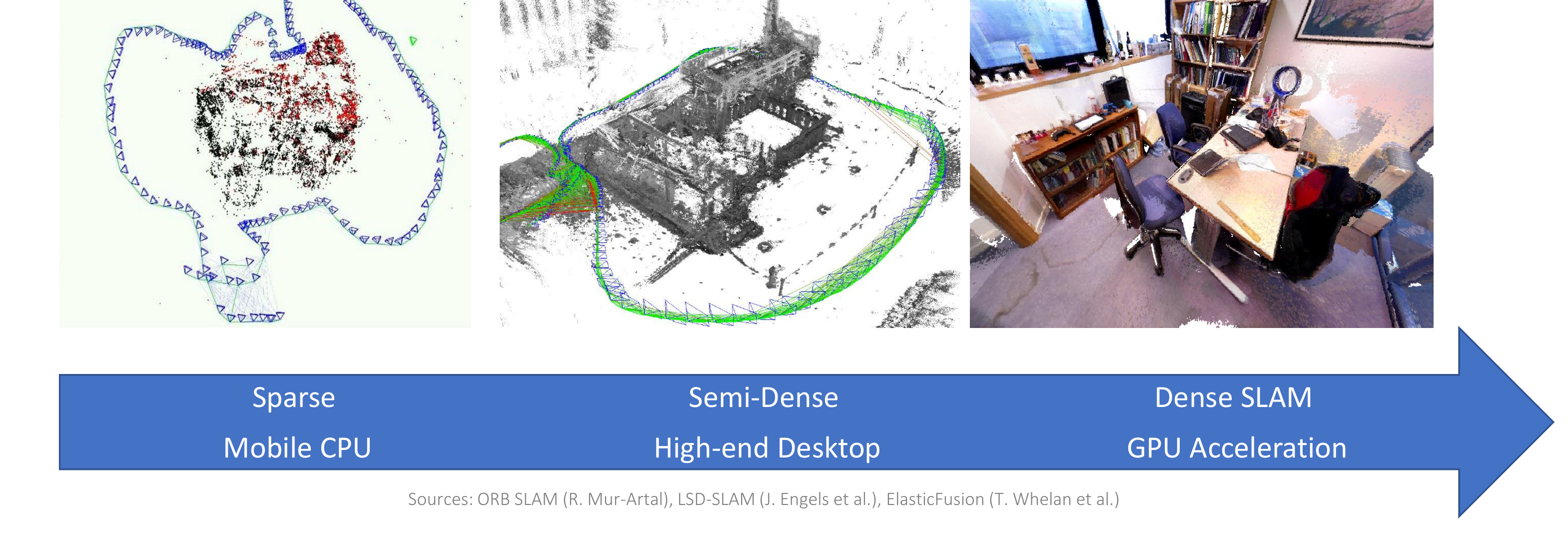}

	\caption{SLAM Continuum from Sparse to Dense}
	\label{arrows}
\end{figure}

Towards addressing the challenges of real-time visual SLAM, the field has gradually split in different approaches, each with their own advantages and disadvantages. A main categorisation is in terms of Sparse to Dense SLAM. Representative examples of these are \cite{mur2015orb, engel14eccv, whelan2015real}, demonstrated in a continuum in Fig.\ref{arrows}. Sparse SLAM uses a smaller set of observations for tracking and maintains a sparse map of the environment consisting of a few points of interest. These approaches exhibit relatively lower computational requirements, but are mainly limited to accurate localisation. 

At the other end of the spectrum, SLAM algorithms categorised as Dense are now able to construct a complete, high quality model of the environment usually as interconnected surfaces. At the same time they are very computationally intensive. In published works the minimum requirement is a high-end multicore desktop CPU for multi-threaded implementations but most have to use GPU acceleration, as for example in the work of Whelan et al.\cite{whelan2015real}, to process all of the available information in real time. Moreover, this trend towards more accurate and advanced but complex SLAM algorithms seems to be continuing at faster pace than advances in computing platforms' raw performance. 

To address this drawback a family of works described as semi-dense SLAM have emerged, e.g. \cite{engel14eccv}. These aim to provide a more dense and information-rich representation compared to sparse methods, while achieving better computational efficiency from processing a subset of high quality observations. However, they are still computationally complex and target desktop-grade multicore CPUs for real-time processing.

\setlength{\textfloatsep}{0.1cm}

Another important distinction is the difference between a full SLAM system and a visual odometry algorithm. Visual odometry focuses on maintaining an accurate position estimate and uses the simplest most efficient form of map possible. On the other hand, full SLAM methods attempt to recover as much of their environment as possible, as well as keep a global, consistent map and enabling loop-closing. Recent solutions, such as SVO \cite {forster2014svo}, can achieve high accuracy tracking using a small set of  high-quality observations. However, much of their efficiency stems from their generation of sparse and local maps which encode significantly less information about the environment. The literature so far has rarely discussed the quality of the maps and re-visiting spaces, with some notable exceptions in specialised works, but in practice it is a crucial aspect for many applications and it significantly affects computational requirements.

There are many examples of emerging applications that require a high level of understanding of their environment that sparse SLAM or visual odometry inherently cannot provide. At the same time, due to safety and robustness requirements, there is often a need for a low processing latency, while most embedded platforms have significant power and weight constraints. These specifications rule out most of the conventional hardware that can perform cutting-edge SLAM in real time. In this context, to close this gap we propose a novel architecture, based on an FPGA-SoC to accelerate semi-dense mapping, targeting state-of-the-art semi-dense SLAM. This accelerator design combines dynamic iteration pipelines and traditional streaming elements to achieve high performance and power efficiency, with a combination of dataflow processing and local on-chip caching to match the unique demands of these algorithms.

Our contributions are twofold. First the design of a scalable and high performance, power efficient specialised accelerator architecture, that can process and update a map in less than \SI{20}{\milli\second} and whose performance can tuned to fit different devices. Second, a system which, when combined with our previous work in \cite{boikos2017high}, forms the first to the best of our knowledge complete SLAM accelerator on FPGAs, pushing the state of the art in performance and quality for SLAM on low-power embedded devices.

\section{Related Work}\label{rela}

Since platforms in the embedded space have significant constraints in power and performance, most embedded visual SLAM implementations focus on sparse SLAM that is adapted towards reducing computational requirements further such as \cite{vincke2012efficient}. The downside to these approaches is that they map a sparse selection of features that reduces the quality of the reconstruction as well as the robustness of tracking in different types of environments.

Another approach towards embedding SLAM has been to design a lightweight, sparse but accurate visual odometry algorithm that can achieve real-time performance on-board an embedded devic\cite {forster2014svo}. This, however, comes with the limitations of sparse odometry algorithms, mentioned in the Introduction. The option of offloading computation to a remote server and reconstructing a dense map there has also been explored \cite{sturm2013dense}. This comes with increased power consumption for the wireless communication, as well as increased latency. It also comes with a reduced area of operation, and very high bandwidth requirements.

Dense SLAM has been advancing rapidly but its requirements in sensors, energy and computation are infeasible for an embedded platform. Works in semi-dense methods such as LSD-SLAM, are more applicable to the embedded space thanks to lower computational complexity and reliance on simpler cameras. LSD-SLAM\cite{engel14eccv} for example, provides a tracking accuracy comparable to other state of the art sparse methods but generates a much denser map that provides more information about the environment. As such, it was selected as the target for the custom accelerator presented in this work.


Recently, there have been attempts in designing custom hardware for SLAM in the embedded space. Suleiman et al. \cite{suleiman2018navion} demonstrated a custom ASIC design for visual-inertial odometry targeting nano-drones. It belongs in the category of sparse odometry and achieves high performance together with power efficiency, realised as a chip printed at 65nm CMOS technology. It enables environment awareness for very lightweight robots, but because of its specialisation it only performs the version of sparse visual-inertial odometry it was designed and cannot be extended to semi-dense or dense SLAM. This is a typical example of an optimised ASIC implementation of an algorithm, which trades flexibility and cost to achieve the highest performance and power efficiency for a specific task.

Most related work on FPGAs in the past has been limited in scope to accelerating selected computation kernels for sparse SLAM such as\cite{weberruss2017fpga}. In contrast, our work targets a more complete implementation of a semi-dense mapping task. Honegger et al. \cite{honegger2014real} proposed a custom board combining an FPGA and a mobile CPU for robotic vision, evaluated by offloading a disparity estimation algorithm (SGM stereo) to the FPGA. Disparity matching with a fixed stereo camera is well-known on FPGAs but is only a pre-processing step needing further processing to be utilized for SLAM. Additionally, their work is focused on a fixed system architecture, providing a one way link with the FPGA between the camera and off-chip memory. In contrast, we target a more flexible system architecture that can allow more fine-grained cooperation between hardware and software.

In our previous work \cite{boikos2017high} we presented an architecture for an FPGA accelerator to provide high-performance tracking for semi-dense SLAM on embedded platforms. However, that work did not provide a solution for mapping, still performed on an embedded CPU at a relatively low performance. This work addresses this and completes the loop of SLAM so that both the demanding and interdependent tasks of tracking and mapping can be offloaded in an efficient way to a reconfigurable platform. The two accelerators are combined to provide much higher performance for the system overall and release the mobile CPU to be used for other tasks.

\section{Mapping Algorithm}\label{mapping_alg}

LSD-SLAM \cite{engel14eccv}, a state-of-the-art semi-dense SLAM algorithm, is the targeted algorithm to accelerate with the proposed coprocessor. The aim of tracking is to recover the camera pose in respect to the world. LSD-SLAM uses the most recent depth observations projected on the current camera frame to optimise directly on the pixel intensity residual. This is expressed as a weighted least squares optimisation, using only the information-rich points in the camera's view. These points are selected based on the intensity gradient in their immediate area. It is then the aim of the mapping algorithm to use the camera pose, estimated from the tracking task, to triangulate points from two views; the current camera frame and the Keyframe, a previous frame in the camera's trajectory stored with its world-to-camera pose along with depth information in a data structure with the same name. That set of depth observations and the selected camera frame on which they project constitutes the current depth map. 

All points with a sufficient gradient successfully matched from Keyframe to camera frame will have a depth value stored in this data structure. Using this information, the mapping algorithm adds a new observation for the points observed for the first time, and performs a filtering update to improve the estimate for points seen in the past. At the end of this process, successfully observed points in space will have an estimated depth and depth variance value stored in the Keyframe. For a more detailed description of the algorithms that constitute LSD-SLAM and the theory behind them one can refer to Engel et al.'s work \cite{Engel_2013_ICCV, engel14eccv}. From this point on, for reasons of brevity the paper will focus on just the information necessary to discuss the proposed custom hardware architecture.

\subsection{Depth Estimation}\label{depth_est}

The aim of this task is to perform an exhaustive search for each high quality point in the Keyframe using its pixel intensity, along a line on the current frame to then be able to estimate its depth. This line is the epipolar line. Geometrically, if we know the relative position and orientation of the camera when two frames were captured, it is proven that a point observed on one camera frame will always project to a line on the plane of the other camera's frame. Two camera frames will not always observe the same point in frame. The line may lie completely outside the frame that a sensor will capture. As such the search is restricted on the intersection of the line and the image frame.

In LSD-SLAM a maximum amount of steps is used to define the search distance. Also, if there is a prior estimation with sufficient confidence, the estimated variance is used to limit the search interval to $d \pm 2\sigma_d$ , where $d$ and $\sigma_d$  denote the mean and standard deviation of the prior hypothesis. At the end of the search for a good match, a sub-pixel accurate localisation is performed for the matching disparity. In  \cite{engel14eccv}, instead of scanning to match a single pixel, a squared error function comparing 5 equidistant points is used to improve accuracy. This approach significantly increases robustness with a small increase in complexity.

In this work, the tasks involved in SLAM were profiled, running as software on an Intel i7-4770 CPU. The results showed that the mapping task was one of the most demanding tasks happening during LSD-SLAM. It consumed 44\% of the computation time spent on SLAM and together with tracking constitutes 85\% of the CPU cycles spent on the SLAM algorithm with the rest spent on pose-graph optimisation and other background tasks. Further testing on the ARM-Cortex A9 of our FPGA board verified the conclusions of the profiling results, with timing tests measuring the mapping task at an average of 530ms per map update.

\section{Architecture}\label{archBIG}

The architecture targets an FPGA-SoC that contains an FPGA fabric and a mobile CPU. The CPU and FPGA can function independently and can operate on the same memory space and both have direct access to a common physical DRAM. There are master memory controllers on the custom hardware for Direct Memory Access (DMA), designed to operate at full-speed bursts for updating the caches before operation or to provide a constant stream of map points for the execution of the algorithm. In addition to the high-speed memory connections, there is a direct slave-to-master connection to the CPU, where the CPU acts as a master. In this manner, the CPU has the high-level control of the coprocessor on the FPGA, and can change its operating parameters and coordinate its operation with the software back end. This part of the system architecture is in a similar philosophy to our work in \cite{boikos2017high}. The way both accelerators were implemented on the FPGA is that they each have exclusive access through an AXI-interconnect to a pair of high performance DMA ports. They share a dedicated DRAM region and the software calls the accelerators to replace the functionality of the software functions. As mentioned in \cite{boikos2017high}, that accelerator has a more fine-grained sharing of computation with the software threads, owing to the iterative, multi-level nature of tracking. In this work, all tasks included in a map update are completely moved to the FPGA and the software only handles the synchronisation of data and tasks.
 
In general, the co-processor architecture is designed to perform  most of the heuristic processing of LSD-SLAM in a streaming fashion. This was chosen to keep compatibility with this state-of-the-art method and maintain the same accuracy and robustness. Nevertheless, in order to increase the performance that is attainable by the proposed custom hardware design, the actual hardware implementation is modified with respect to the original software implementation. For instance, a number of values, such as the maximum gradient in a neighbourhood were more efficiently calculated on the fly than pre-computed as done in software. Additionally, most of the functions in the algorithm are combined in one streaming pipeline utilizing buffers to overlap computation, as this avoids redundant memory traffic and significantly improves performance and power efficiency. Finally, to remain faithful to the algorithm, most of the computation happens on floating-point as in the original implementation. There are variables where high-dynamic range or multiple divisions make the floating-point implementation necessary for accuracy or performance reasons. For the rest of the units the conversion to fixed-point arithmetic was not straightforward but requires careful analysis. However the principles behind choosing the most suitable arithmetic representation are well known in the field of custom and reconfigurable computing. As such, we chose first to focus on developing the most suitable architecture, presented in this work, leaving custom-precision representation as future work after the system and the microarchitecture were fixed.

\subsection{High-level Functionality and Algorithm Mapping}\label{archandfunc}
\begin{figure}[h] 
	\centering
	\includegraphics[trim={2cm 4.8cm 1cm 2.5cm},clip ,width=1\linewidth]{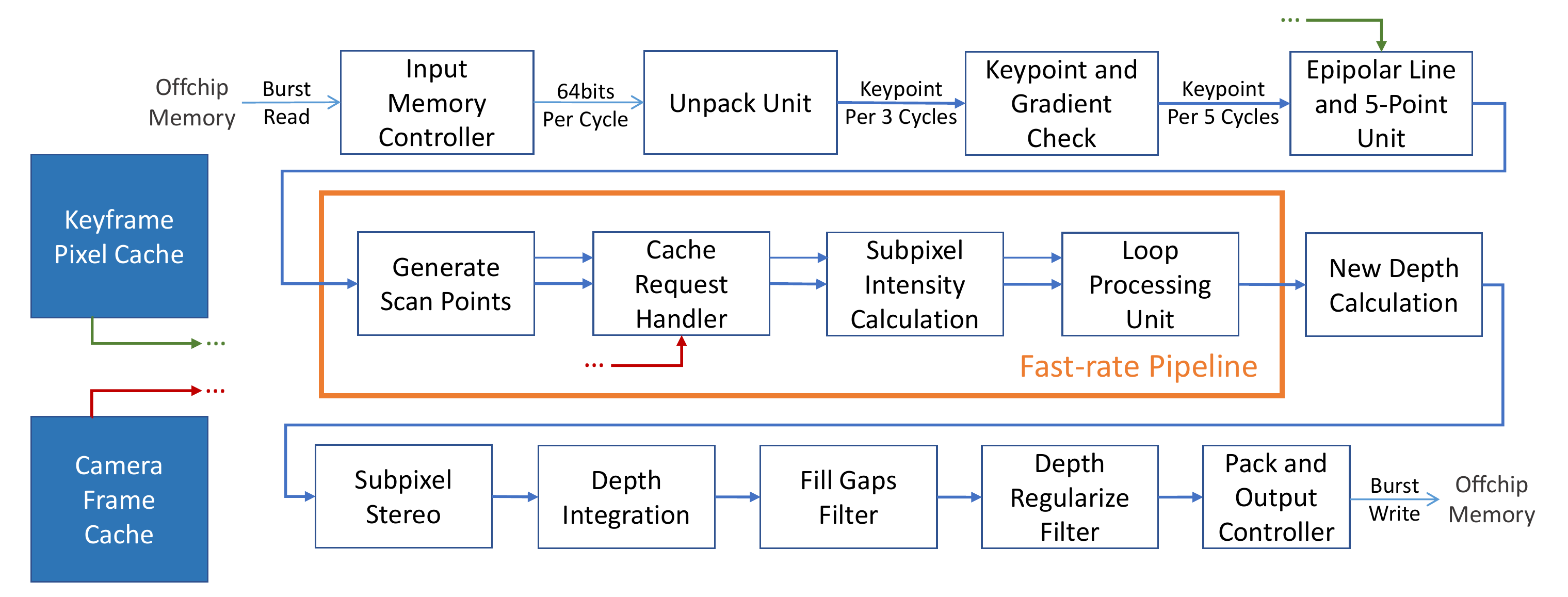}
	
	
	\caption{Block diagram of the coprocessor architecture}

	\label{arch}
\end{figure}

Figure~\ref{arch} contains a high level view of this architecture omitting some connections for clarity. The first step is the update of the caches if necessary. Then as input the architecture receives all the points of the Keyframe sequentially as described in Section \ref{mapping_alg}, and its output is the final state of the updated Keyframe data structure, again output sequentially. The first two units ensure a fast and consistent stream of Keyframe points. The `Input Memory Controller' performs full-speed burst reads from the off-chip memory, that are then buffered and streamed as Keypoints from the `Unpack Unit' to the rest of the pipeline.

As the Keypoints stream in, the `Keypoint and Gradient Check' unit is responsible for calculating on the fly the max gradient in a neighbourhood of the pixel. Based on the gradient threshold for the area and the pixel's confidence rating, the Keypoint's fitness is calculated as a candidate to try to map. It is then forwarded to the `Epipolar Line and 5-Point Unit' that is responsible for calculating the scan range, center and steps. Next, a check of the robustness of the search is performed, including if it is inside the frame's limits. If all the checks are valid, this information is forwarded to the fast-rate pipeline. If it fails, the map point is still forwarded to be used for later processing such as filtering, followed with flags to mark this decision and the reason for failure.

In the fast-rate pipeline, as shown in Fig.~\ref{arch}, the thinner lines correspond to the map point together with its metadata being forwarded. The main operation of these units has to do with the scan and best match selection on the epipolar line. The information pertaining to this scan is passed between them at a faster rate as long as the scan is going on for one single point, indicated by the thicker lines in the center of the units. In this faster rate pipeline, the `Generate Scan Points' unit supplies a steady stream of pixel locations to be fetched from the cache unit, according to the calculations in the Epipolar line unit. The `Cache Request Handler' fetches these pixels from the caches and forwards them to the `Subpixel Intensity Calculation' unit where linear interpolation is performed in a neighbourhood of 4 pixels around the floating point coordinates. 

All these streams are passed on to the `Loop Processing Unit' (LPU) that performs the core of the scanning algorithm. It reconstructs the pattern of 5 pixels we are looking for and performs the scan steps to find the position with the minimum sum of squared errors. It keeps the best match and second best match and additional information regarding the search. This includes the steps performed, the distance of the search and the match error.  After a scan is completed, this is forwarded to the `New depth Calculation' unit. This calculates a new depth and depth variance value based on the results of the LPU, which the next unit `Subpixel Stereo' can further refine if the conditions are right.

Finally, the `Depth Integration' and the Filter units. The first is responsible for putting all the information together for each map point, and the filter units perform regularization operations. The first one, if it finds sufficient confidence in a window around a pixel without an observation, fills it with a weighted average of its valid neighbours. The second filter calculates a smoothed value for the depth and variance of valid map points, stored separately to the actual depth, again operating on a sliding window around a center pixel. Here row buffers allowed region of interest processing, without breaking the streaming interface of the filter units. After the processing and filtering finishes, we reverse the operations at the input in a pack-and-output unit that streams it out to the off-chip DRAM with burst write transactions.

\subsection{Multi-rate dataflow operation} \label{multirate_dataflow}

Semi-dense SLAM is characterised by a large amount of data that needs to be processed. For a map of size $640\times 480$ there are 7.37MBytes for the depth map representation. That is in addition to the actual frame size of 307kBytes. To put that into perspective, in order to process 60 frames/sec as they come from a camera, and extract depth information for all of them the total time between captured frames would be less than 17ms, but that amount of data requires approximately 8-10ms just to be read from memory with the typical memory bandwidth available on off-the-shelf FPGA-SoCs. To keep up with that time it would be necessary to process one map point every 6 cycles on average. A straightforward implementation trying to perform all necessary epipolar line scan steps inside this time would provide a large, underutilized design, with a high power consumption. 

Alternatively certain properties of semi-dense SLAM can be leveraged to design a much more efficient solution. An epipolar line scan often is not required when the point does not currently contain a valid observation or is not visible in the current frame. Moreover, in confident observations, it can be safely reduced to the region $d \pm 2\sigma_d$, as described in Section \ref{depth_est}. The designed coprocessor takes advantage of the pattern and frequency of the aforementioned cases by utilizing fully pipelined units, each designed to efficiently execute a part of the computation of the entire algorithm, as discussed in Section \ref{archandfunc}. 

We found the most efficient design to be self-contained, deeply-pipelined hardware blocks that perform different types of operations by re-using math units, clocked at a synchronous rate, while logic changes the operation path. This way we overlapped different parts of the algorithm in the same hardware units, and designed everything with the principle of data always moving forward. The pipelines contain multiple math units for multiplication, addition and division, and logic and multiplexers shift the structure of the unit as necessary. This way they can change from an initialization phase, to operating on points, to scanning across the frame cache, depending on the unit, or skipping a scan and forwarding metadata to the next unit in the pipeline. 

The units were also designed to operate at different rates, with fast-rate processing units in the middle to perform epipolar scans, find the best match and perform the depth estimation, and more relaxed processing at most of the input and output stages. The cache accessing was normalised to one access window per cycle, with buffering and control allowing a very simple and high efficient cache controller to serve different kinds of requests from other units. The units are connected to each other through large streaming FIFO buffers that allow communication to happen asynchronously, and hide a lot of the latency that would arise from the variable processing rate design. In this way we offer a much higher performance level, but use a fraction of the resources of a pipelined statically allocated for the worst processing load.

As shown in figure~\ref{arch}, the units in the fast-rate pipeline operate at a faster rate. When an epipolar line scan and depth update is necessary, they perform some initialisation steps at a rate of one step per cycle. Otherwise, if there is a point that does not require a depth update, they directly forward that point's metadata to the next unit in a single cycle. Finally they use most of their resources in a normal operation to perform one scan step per cycle. Their accesses to cache are pre-computed and pre-fetched at the `generate scan points unit' so they perform the necessary operations directly on incoming data.

By reducing the amount of multiplier, divider and accumulation units built in each unit, and time sharing them more aggressively for different operations, we can increase the amount of cycles necessary for a scan but with an almost linear decrease in resources for that unit. The most efficient designs must have units in the fast-rate pipeline match with each other, as otherwise the slowest one would dictate rate of processing, leaving unused resources in the rest. In a similar fashion the rate of processing for the units before the fast-rate pipeline should be tuned as one number, and the same or slightly slower processing rate should be targeted for the units after the fast-rate pipeline. The resulting architecture is tunable in terms of its performance and resources, allowing it to scale to different FPGA devices and resource budgets. In Section \ref{eval} we show different example design points achieved by changing the target processing rates as described previously.

\subsection{Performance Analysis} \label{perf}

To explore the optimal hardware rates described in Section \ref{multirate_dataflow}, and verify that our design assumptions hold when running with real-world datasets, monitoring instrumentation was added in the software version of LSD-SLAM and it was executed for the entire duration of real datasets. Firstly, we collected statistics regarding the average processing load that is expected for each iteration of a map update, and then studied the distribution and extrema of these samples.

\begin{figure}[h] 
	\centering
	\includegraphics[trim={3cm 2cm 3cm 2.1cm},clip ,width=1\linewidth]{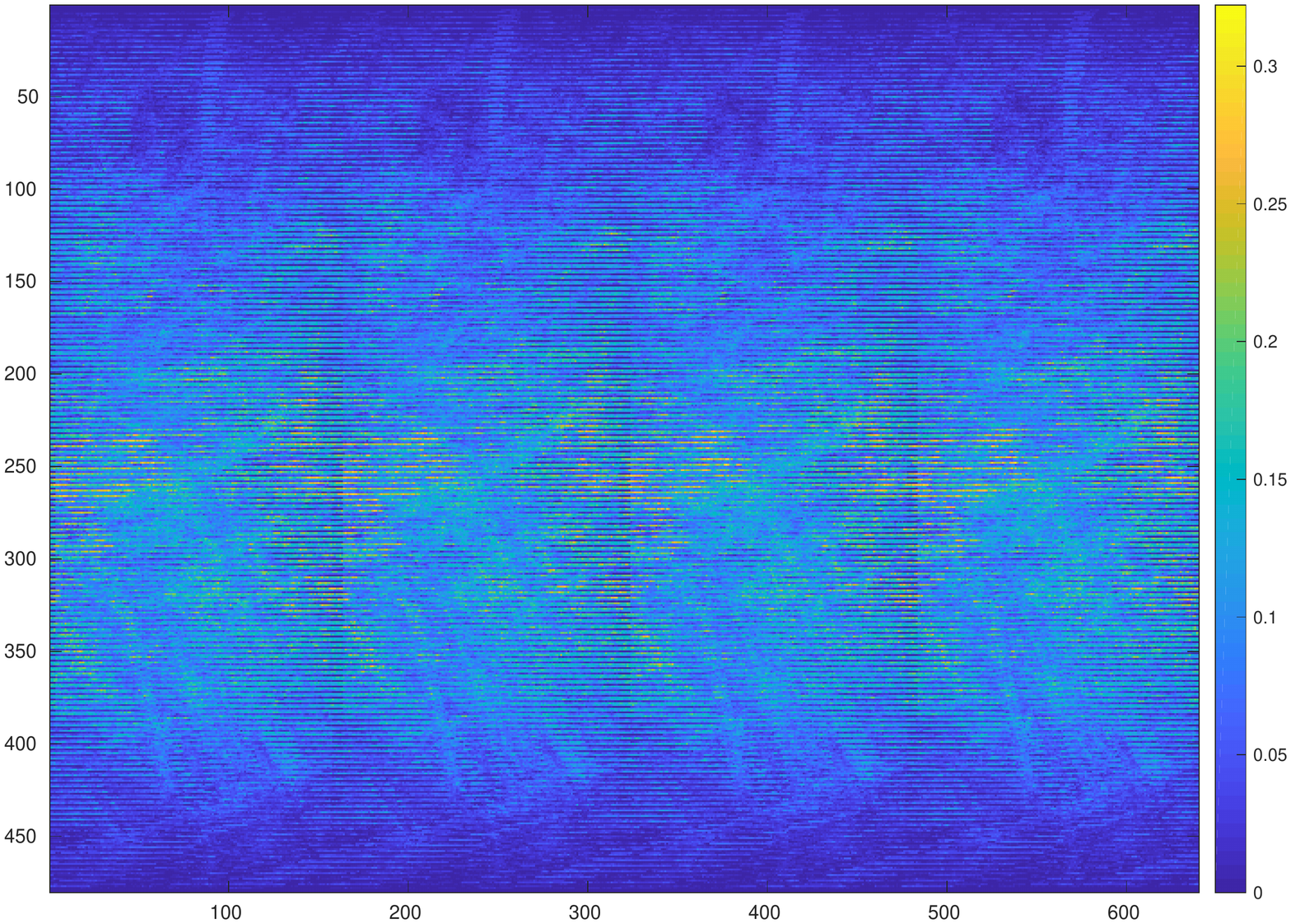}
	
	
	\caption{Heatmap representing scan frequency for mapped Keyframe points}
	\label{heatmap}
\end{figure}

A subset of this data is visualised on the heatmap of Fig.~\ref{heatmap}. Its dimensions are equal to the Keyframe image size. Each cell corresponds to a pixel of the Keyframe, and the colour indicates, the frequency (0-1) with which that location will contain a valid point requiring an epipolar line scan. We can see that the frequency for any point is consistently lower than the 30\% mark. Further testing for peak loads across a dataset revealed that the average amount of points per line that require scanning peaks around the center of the image at a frequency averaging 18\%. By looking for extrema we discovered some outlier cases, which however were usually less than 1-2\% of the frames processed. Those have to do with special cases consisting of initialisation steps or very sharp motions. However, the worst case scenario will always have an upper bound, and have a linear relationship with the fast-rate pipeline processing rate and the processing load per frame. Thus, it can be predicted and designed against.

In the implemented unit, a processing rate of one scan/interpolation per cycle was chosen for the fast-rate pipeline and a processing rate of one target point per 5 cycles in all the other units. For the datasets tested, this relationship of 5-to-1 was a good ratio for the processing rates of the pipelines, with the majority of frames not filling the buffers completely. The average case for one epipolar line scan was calculated at 11 steps for the presented design. Given the results in Fig.~\ref{heatmap}, if 25\% of the points in a line require an epipolar scan, and the other 75\% are skipped in one cycle, the total latency per row would be 2240 cycles, or 3.5 cycles per point for an average of 11 scan steps, which leaves a good margin of safety for heavier loads.

In datasets tested, 98\% of frames were within one millisecond of the target processing time and more than 99\% were within two. There were some outlier cases with a performance drop of up to 30\%, from 16.3ms to 21 ms. For example in the Machine hall dataset, one of the two depicted in the Evaluation Section, out of 3268 map updates only 19 were around the \SI{20}{\milli\second} mark, with a maximum recorded value of 20.9ms. However that is considered acceptable in this application for two reasons. Firstly, the application can dismiss more than one dropped map update out of 100 and can handle a lower mapping rate than the one we targeted. Moreover, in actual tests the software version on both tested platforms had a worse behaviour in outliers with an increase of almost 200\% in the processing time for some cases. 

Secondly, the proposed architecture is tunable and can be changed to adapt to different application requirements. One can increase the capabilities of the fast-rate pipeline to have the system guarantee a very small performance degradation even in outlier cases at the cost of some underutilized resources. Alternatively, if the application allows, one can go the other way and under-provision the fast-rate pipeline to target a more resource-and-power efficient system, by allowing some degradation of a few percentage points in more cluttered scenes. In Fig.\ref{resources}, we can see the scaling to target different performance points. The 32.5 fps and 42.5 fps are examples of a design point where an extra cost in resources guarantees a lower maximum latency, and therefore a higher target performance.

\section{Implementation and Evaluation}\label{eval}

\begin{figure}[!h] 
	\centering
	\includegraphics[trim={1.8cm 2.2cm 2cm 3.4cm},clip ,width=1\linewidth]{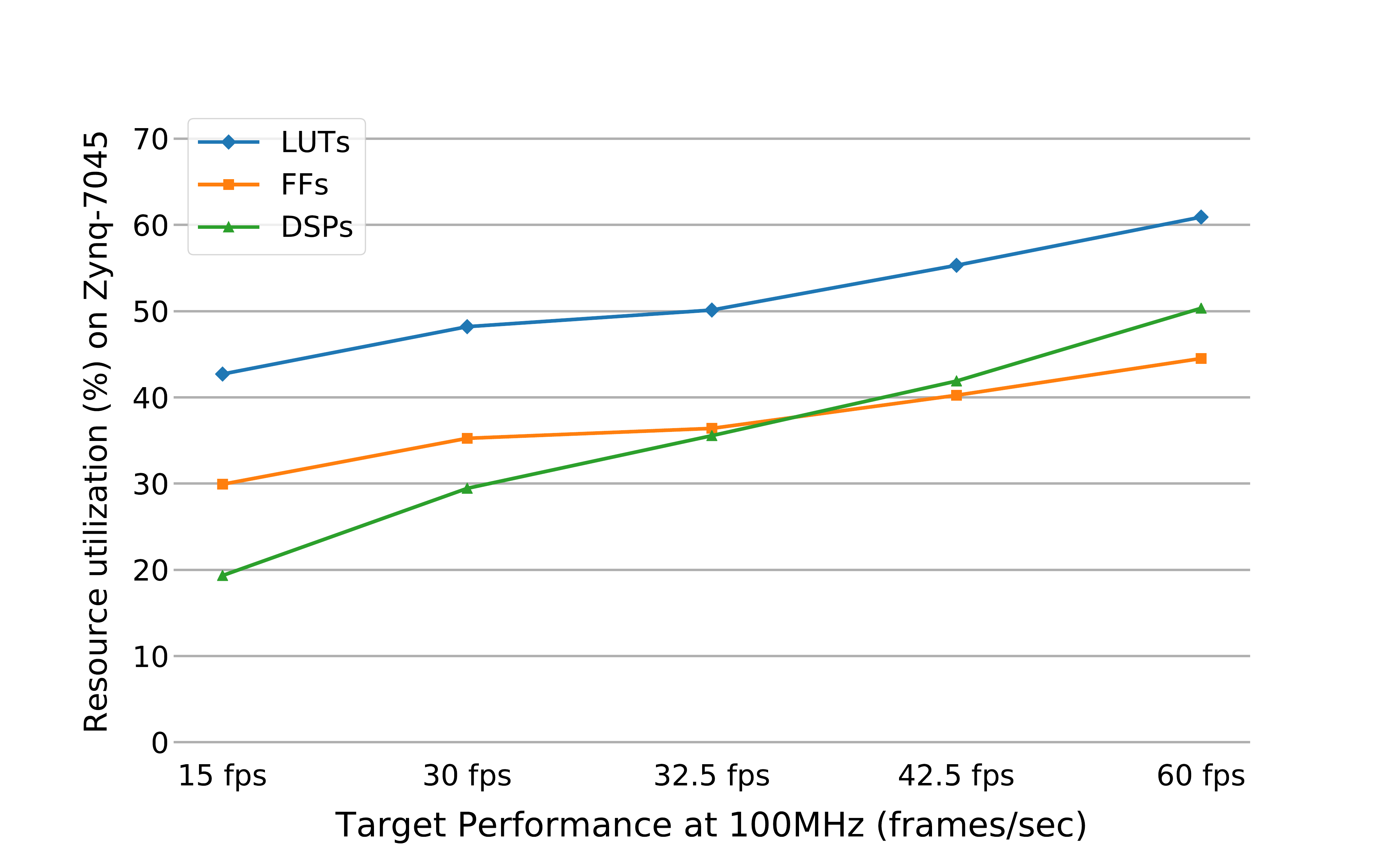}
	
	
	\caption{Resource scaling with architectural tuning targeting 100MHz}
	\label{resources}
\end{figure}

Fig.\ref{resources} demonstrates scaling from a lower perfomance point that can target one of the smaller Zynq devices the Zynq-7020, to the 60fps design point, which was selected to allow a second accelerator to fit alongside in the larger ZC706 board (Zynq-7045). We can see that some resources such as the DSPs, ubiquitous in most math units, scale almost linearly with the target performance, followed by the LUTs, while Flip-Flops have a standard offset cost owing to their extensive use in I/O and memory access units which were not part of the tuning process. The architecture described in the previous Section is designed to be platform agnostic and optimised on resource usage. Nevertheless, the use of Vivado HLS tools drove a number of implementation decisions in order to develop and test the IP on the target FPGA-SoC, leading to certain overheads\footnote{For example, since the tool always rounds up memory size to the next power of two for BRAM utilization, we elected to partition in two dimensions cyclically by a factor of 5, a non-power of 2 factor. This significantly reduced the memory overheads, fitting eventually both accelerators on the same device in the highest performance configuration, at the cost of increased DSP and LUT usage.}. 

For evaluation, the design was synthesized and placed-and-routed with Vivado HLS and Vivado Design Suite (v[2018.2]), targeting a Xilinx Zynq ZC706 board and run and tested on the same board. For the parameters described in Section \ref{archBIG}, timing was met for the coprocessor at 125 MHz. The resource usage for that result, post-implementation, is described on Table \ref{tab}. Combined with our design from \cite{boikos2017high} executing on the same reconfigurable fabric, the accelerators were successfully tested working side-by-side, setting the target frequency to 100 MHz, replacing key functions in the software implementation running on the mobile CPU.

	\begin{table}[h]
		
		\centering

		\begin{tabularx}{0.94\textwidth}{XXXX}
			Resource & This work & With \cite{boikos2017high}  & Available on Z-7045\\ \hline 
			
			LUT      & 151,674	& 	184,993	& 218,600          \\
			LUTRAM   & 12,242	&    15,317 	& 70,400           \\
			FF       & 213,761	&  	256,665	& 437,200          \\
			BRAM     & 958		&    1089  	& 1090             \\
			DSP      & 594		&    718   & 900       \\
			\hfill     
		\end{tabularx}

		\caption{Resources post-implementation}
		\label{tab}
	\end{table}
	
\begin{figure}[] 
	\centering
	\includegraphics[trim={2.2cm 3.8cm 6.2cm 4.6cm},clip ,width=1\linewidth]{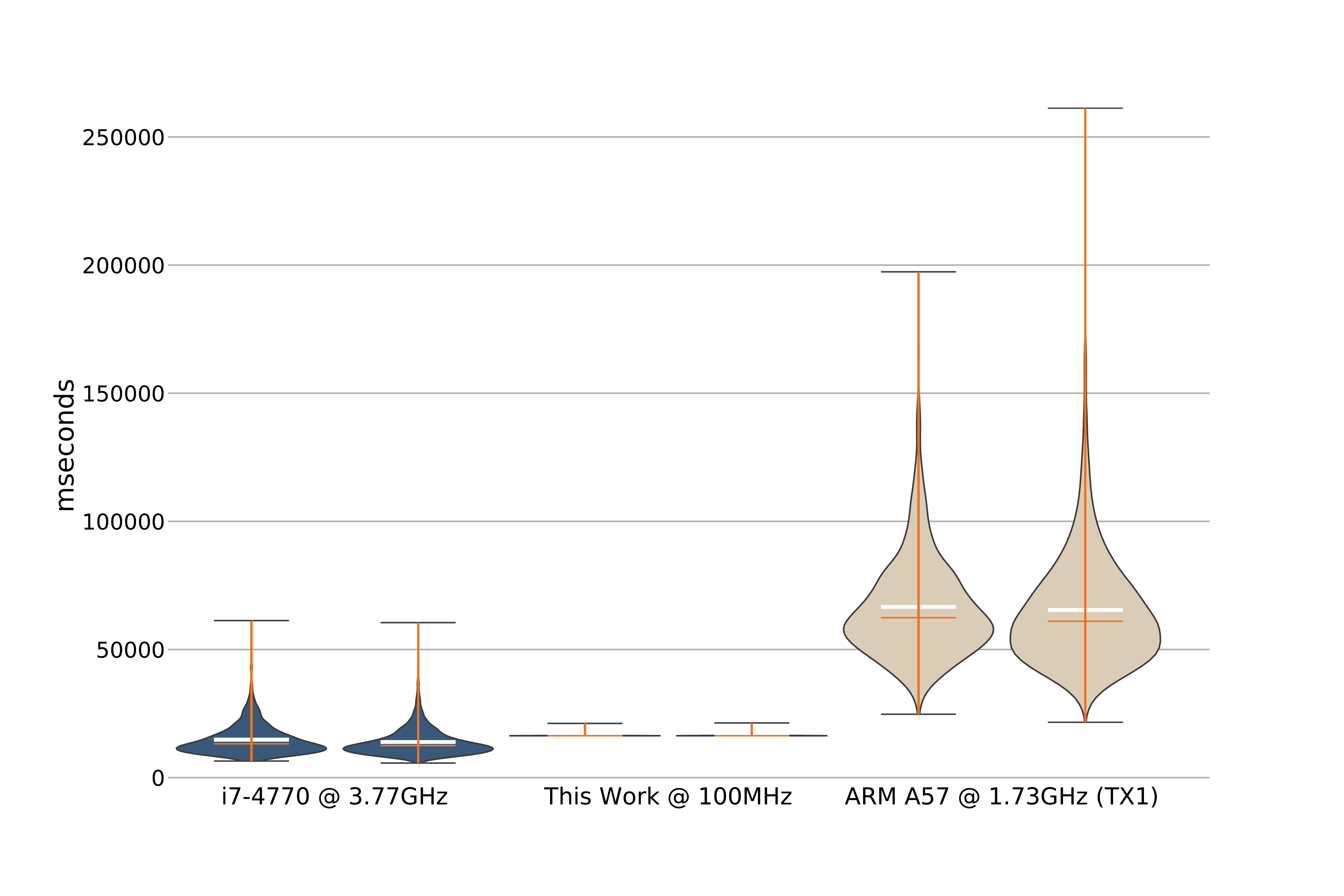}

	
	\caption{Mapping latency in ms - Different Platforms on two Datasets }
	\label{violins}
\end{figure}

On Fig.\ref{violins}, we can see the mapping performance (total processing time for a map update step) on three high-end platforms across two separate datasets\footnote{These were the Room and Machine Hall trajectory from TUM's website: https://vision.in.tum.de/research/vslam/lsdslam}. The colour corresponds to the platforms, an Intel i7-4770, our accelerator implemented on a Zynq-7045 and the Cortex-A57 on a Tegra TX1. The width of the shape corresponds to the density of observations around a particular value of milliseconds, similar to a sideways kernel density plot. The thicker, white line in the middle corresponds to the mean value of the observations, while the thinner, orange one to the median value. Finally the lines at the top and bottom are the minimum and maximum values observed. The figure demonstrates the variability of this processing load on general purpose hardware, and how robust this accelerator is to these delays, appearing almost flat since most observations were very close to the ideal value of approximately \SI{16.2}{\milli\second} at \SI{100}{\mega\hertz}.

In addition to performance, power consumption was measured for each platform at the wall, including board and power supply losses. Static and dynamic power are separated to demonstrate the chip power contribution at full load. The measurement is accurate to $\pm 0.5$ watts, an accuracy sufficient to reach some conclusions for these different platforms. In the case of our accelerator we can estimate approximately 1-2 watts of the static power draw to be due to the FPGA. Testing power draw with an empty bitstream on the FPGA showed a decrease in static power of approximately 2 watts adding merit to this. We achieve a performance on par with the high-end desktop CPU, but for an order of magnitude less power consumed at the FPGA fabric.

\begin{figure}[h] 
		\centering
		\includegraphics[trim={1.6cm 2cm 1.6cm 3cm},clip ,width=1\linewidth]{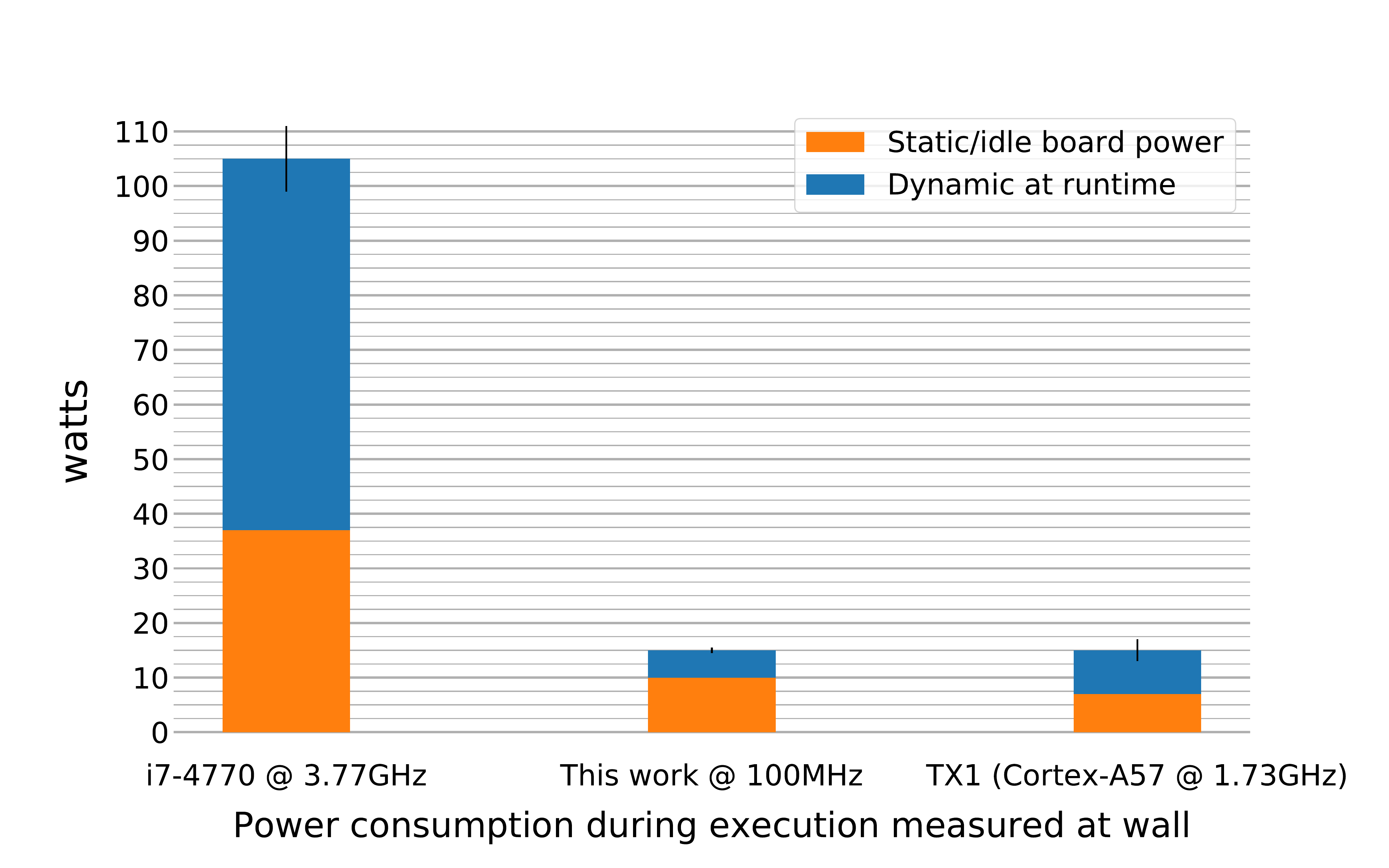}
	
		\caption{Power consumption of the devices tested. }
		\vspace{0.1cm}
		\label{power}
	\end{figure}

We can see that the FGPA development board is at a similar power level at full load with the Tegra TX1, but with more than a 4x increase in performance on average for our accelerator design. We estimate the total power of mobile CPU + FPGA fabric on the Zynq-7045 at 6.5 watts, using the estimator on Vivado post-implementation, combined with the results shown on Fig.~\ref{power}. Static is high since it includes several unnecessary peripheral devices on the FPGA board such as a second DDR memory. On the Tegra, the GPU was set to run at the lowest clock setting so that the power measurements would reflect mainly the CPU's behaviour.

{\renewcommand{\arraystretch}{1.3}%
	
	The aim of the accelerator, together with our previous work \cite{boikos2017high}, was to provide a complete acceleration solution for LSD-SLAM, a state-of-the-art semi-dense SLAM method. The two designed architectures both achieve real-time performance, evaluated running LSD-SLAM with a pre-recorded dataset utilizing the two accelerators, with a board power draw at the wall of \SI{15}{\watt}. So far we have compared the performance of the accelerator to that of the software implementation executing in an embedded and a desktop-grade CPU. In Table~\ref{slam_examples} we collect some representative examples of the current state of the art both in SLAM algorithms as well as typical embedded solutions. 
	
	\vspace{-0.2cm}
	\begin{table}[h]
		\centering
		
		\begin{tabular}{lccccccr}
			Work    	  & Type          & Hardware Plat.    & Density & Close-loop  & Inertial & Typical Power \\
			\hline
			
			\cite{mur2015orb} 	& SLAM     & Laptop CPU           & Sparse         & \checkmark  &            & 38-47W   \\
			
			\cite{engel14eccv}	 & SLAM     & Laptop CPU           & Semi-dense     & \checkmark &            & 40-50W   \\
			\cite{whelan2015real}	& SLAM      & GPU Accelerated      & Dense          & \checkmark &           & 170-250W \\
			Ours   & SLAM & FPGA SoC & Semi-dense     & \checkmark &           & 6-7W     \\
			\cite{leutenegger2015keyframe} & SLAM   & Laptop CPU           & Sparse         & \checkmark & \checkmark &  30-50W   \\ 
			\cite{forster2014svo}  	 & Odometry   & Laptop/Jetson-Tx1 & Sparse         &           &           & 30-40W/10-15W          \\
			\cite{weberruss2017fpga}  & Kernel Acc.    & FPGA & Sparse        &           &                     & 5.3W     \\
			\cite{suleiman2018navion}   	& Odometry         & ASIC - 65nm CMOS     & Sparse         &           & \checkmark  & 2-24mW   \\
			
			\hline      \\
		\end{tabular}
		\caption{State-of-the-art SLAM examples. Compiled with a focus on features and characteristics of different solutions to demonstrate the breadth of the field}
		\label{slam_examples}
		
	\end{table}
	\vspace{-0.8cm}
	
	The table is not meant to be exhaustive or rank the works. Instead, it was compiled to focus on the characteristics of different solutions and provide an overview of different software and hardware approaches to SLAM and their power characteristics\footnote{The power figures were often not mentioned in works, or measured with varying methods. Thus, in the interest of providing a qualitative view, we include a typical expected power for the chip/platform mentioned in the publications (e.g. nVidia 680GTX, Jetson TX1, Intel i7-4700MQ etc.). For our work, we report the estimated chip power instead of the board power to be in line with other papers.} The key takeaway is the gap between fast but sparse odometry with no large-scale capabilities or loop-closure on embedded systems and accurate, complex and dense solutions occupying different positions on the algorithmic landscape but requiring high-end hardware for real-time operation.

\section{Conclusions}

Our findings were that the most efficient designs for the target application combine features that include a high-bandwidth streaming interface to common memory and local caching of the region of interest or if possible the entire image frame processed. Dealing with the complex control-flow of these algorithms we found the most efficient  choice to be multi-rate, multi-modal units, separated by buffers. We also found the most efficient and high performance choice to be a pipeline design that follows the dataflow paradigm, trying to move every data point through once. To have the most efficient design we separated the memory accesses from the actual computation, and carried unit control parameters as metadata along the processing path leading to more efficient designs. 

In conclusion, this work proposes an FPGA-based architecture that achieves the required performance to run high quality state-of-the-art semi-dense SLAM with high-end desktop performance at the power level of an embedded device. It  has good scalability and is parametrised to address various SLAM specifications and target different FPGA-SoC devices, demonstrated by successfully running alongside the accelerator from \cite{boikos2017high} to provide cutting-edge performance.

\section*{Acknowledgments}
The support of the EPSRC Centre for Doctoral Training in High Performance Embedded and Distributed Systems  (HiPEDS, Grant Reference EP/L016796/1) is gratefully acknowledged.

%
%
\bibliographystyle{splncs04}
\bibliography{bibli}

\end{document}